\documentclass[runningheads,a4paper]{llncs}
\usepackage[T1]{fontenc}
\usepackage{wrapfig}
\usepackage{graphicx}
\usepackage{amsmath}
\usepackage{multirow}
\usepackage{multicol}
\usepackage{array,booktabs}
\usepackage[skins,theorems]{tcolorbox}
\usepackage{hyperref}
\usepackage{tabularx}
\usepackage{xcolor}
\usepackage{gensymb}
\usepackage{placeins} 
\usepackage{setspace}
\newcommand{\yash}[1]{}
\newcommand{\young}[1]{}
\newcommand{\kar}[1]{}
\newcommand{\RED}[1]{#1}

\newcommand{\qq}[1]{\(\text{NightHawk}\)}
\begin{document}
\title{Active Illumination Control in Low-Light Environments using \qq{}}

%
%


\author{Yash Turkar \and
Youngjin Kim  \and
Karthik Dantu
}

\authorrunning{Turkar et al.}
%
\institute{University at Buffalo, State University of New York \\
\href{https://droneslab.github.io/NH/}{droneslab.github.io/NH} \\
\{yashturk, ykim35, kdantu\}@buffalo.edu}
%
\maketitle              
\vspace{-0.7cm}
\begin{abstract}

Subterranean environments such as culverts present significant challenges to robot vision due to dim lighting and lack of distinctive features. Although onboard illumination can help, it introduces issues such as specular reflections, overexposure, and increased power consumption. We propose \qq{} \footnote{This project was partially funded by NSF \#1846320 and a gift from MOOG Inc.}, a framework that combines active illumination with exposure control to optimize image quality in these settings. \qq{} formulates an online Bayesian optimization problem to determine the best light intensity and exposure-time for a given scene. We propose a novel feature detector-based metric to quantify image utility and use it as the cost function for the optimizer. We built \qq{} as an event-triggered recursive optimization pipeline and deployed it on a legged robot navigating a culvert beneath the Erie Canal. Results from field experiments demonstrate improvements in feature detection and matching by 47-197\% enabling more reliable visual estimation in challenging lighting conditions.


\end{abstract}
%
%
%


\vspace{-0.8cm}
\begin{figure}
    \centering
    \includegraphics[width=\linewidth, trim=1.8cm 1.5cm 1.8cm 1.0cm, clip]{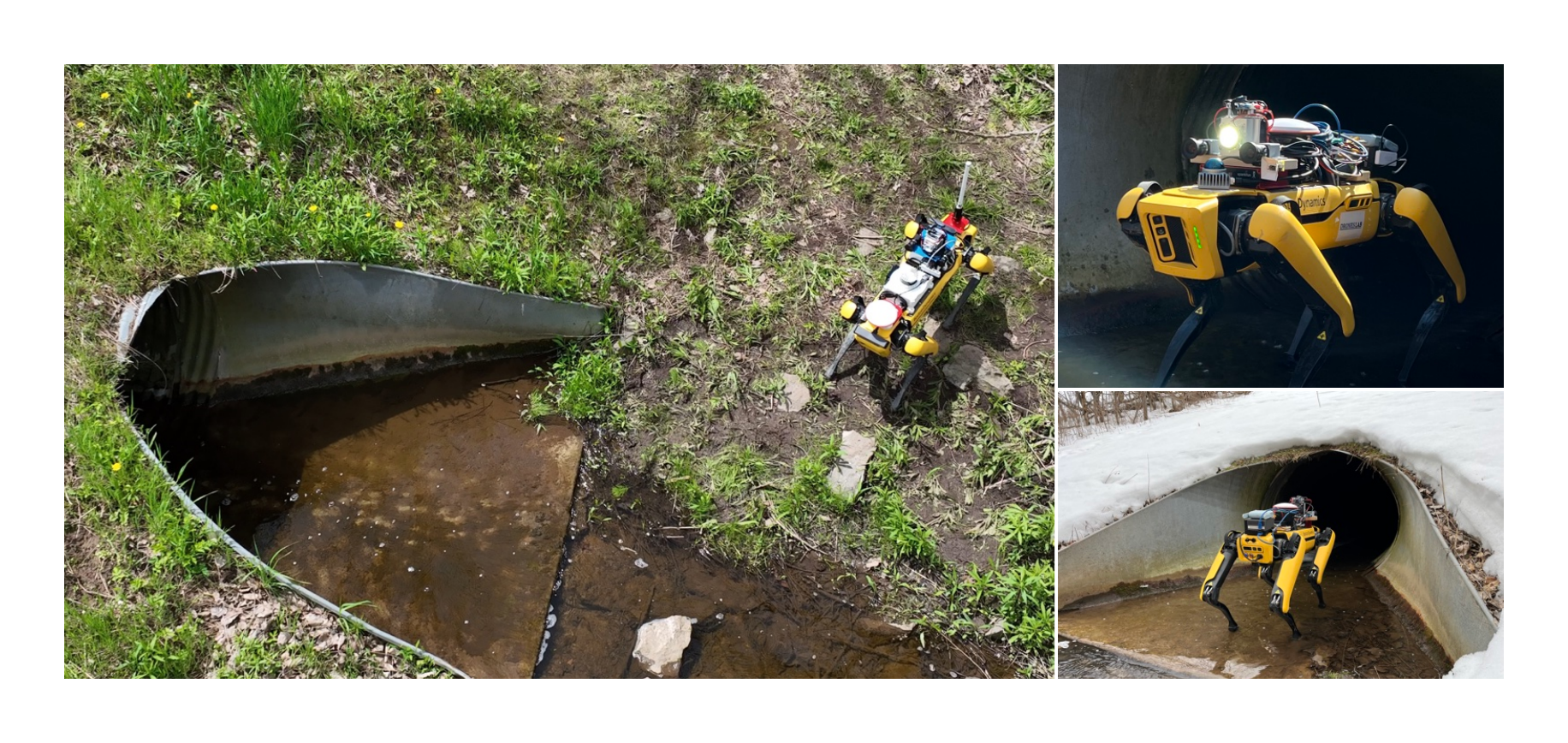}
    
    \caption{Active illumination and exposure control on Boston Dynamics Spot, inspecting culverts under the Erie Canal in Medina, NY. (Left) Spot approaches entrance of culvert 110, a 66m long, subterranean environment with extremely low ambient illumination. (Right) Spot entering and exiting the culvert equipped with \qq{}} 
    \label{fig:intro}
\end{figure}
\vspace{-0.8cm}

\section{Introduction}


\RED{Environment illuminance plays a critical role in the performance of robot perception algorithms, many of which rely heavily on feature detection and matching. Illuminance directly influences image brightness and camera exposure, ultimately affecting image utility. This impact is especially pronounced in scenarios where lighting conditions are variable, such as indoor-outdoor transitions, shadowed areas, or environments with dynamic or low lighting.
Standard vision sensors, including monocular, RGB-D, and stereo cameras, rely on autoexposure (AE) to adjust settings like exposure time (shutter speed), gain (ISO), and aperture. While AE performs well in typical (well-lit) scenarios, it struggles in extreme environments. The primary limitation of AE stems from its objective: maintaining a mean pixel intensity around 50\%, aiming for neither over nor under exposure. However, this approach may not be optimal for robot vision tasks. The goal in these applications is to reliably detect and match stable features, which a neutrally exposed image doesn't necessarily guarantee.  


Several exposure control methods have been proposed~\cite{zhang_image_2024},~\cite{han_camera_2023},~\cite{zhang_efficient_2025}, \cite{kim_exposure_2018}, \cite{mehta_gradient-based_2020}, \cite{lee_learning_2024}, \cite{gomez-ojeda_learning-based_2018}. \cite{liu_learning-based_2020}, controls camera exposure-time  while \cite{tomasi_learned_2021} controls both exposure-time and gain. Some of these methods propose image utility metrics that aim to quantify the quality of images from a feature detection and matching perspective like \(M_{shim}\)~\cite{shim_auto-adjusting_2014}, \(M_{softperc}\)~\cite{zhang_active_2017} and NEWG~\cite{kim_proactive_2020}.  These metrics typically revolve around using image gradients, as most feature detectors exploit gradients for keypoint detection.
These exposure control methods adjust camera parameters such as shutter speed and gain, but often fall short in challenging low-light or varying-light conditions.  Insufficient scene radiance can necessitate excessively long exposures or high gains leading to reduced frame-rates and increased noise. Integrating an onboard light source offers a promising solution by augmenting scene illumination. However, naive control of onboard lighting can introduce undesirable artifacts such as specular reflections \cite{ebadi_present_2024} and overexposure, while also consuming significant robot power. Careful tuning and adjustment of light intensity is essential to mitigate these drawbacks. }

\RED{Our motivation stems from our efforts to inspect culverts beneath the Erie Canal in western New York. These culverts are long pipes with a 1-meter diameter that extend across the canal. As shown in \autoref{fig:intro}, they are dimly lit with extreme light variations at the entrances. They are characterized by repeating textures and features, posing a significant challenge for visual estimation. }


\RED{
We address these challenges with \qq{}, a novel framework that combines active illumination with exposure control. We also present a new image utility metric (\(M_{feat}\)) which leverages learning-based feature detectors to assess image quality and demonstrates strong correlation with feature matching performance in low-light settings. This metric serves as the cost function for our online Bayesian optimization process enabling us to determine optimal external light intensity and camera attribute values. Our key contributions are:
\begin{enumerate}
   \item Novel image utility metric (\(M_{feat}\)) based on a modern learning-based feature detector that effectively quantifies feature performance
   \item An active illumination and exposure control framework (\qq{})  which uses online Bayesian optimization to find optimal external-light intensity (\(P\)) and exposure-time (\(\Delta t\))
   \item Experimental validation in a challenging subterranean environment—a 66-meter-long, 1-meter-diameter culvert beneath the Erie Canal—to demonstrate enhanced feature matching performance.
\end{enumerate}}
\section{\qq{} Design}
\label{nighthawk-section}
\qq{} is an external light and camera exposure-time control algorithm that uses event-triggered Bayesian Optimization to provide optimal lighting and camera configuration.  This section describes the image utility metric and illumination control strategy.




\subsection{Image Utility}
Effective exposure control requires a reliable mechanism to quantify image quality. Conventional auto-exposure (AE) mechanisms rely on irradiance aimed to maintain a mean intensity of 50\% or 128 (8-bit). Prior approaches proposed by \cite{shim_auto-adjusting_2014}, \cite{zhang_active_2017}, \cite{kim_proactive_2020} control exposure by using image gradient-based utility metrics. Recently, there have been numerous learning-based feature detectors (e.g., SiLK\cite{gleize_silk_2023}, R2D2\cite{revaud_r2d2_2019}) that are trained in a self-supervised manner to estimate probabilities of being "interesting" per pixel. Further, detectors such as R2D2 explicitly output repeatability and reliability tensors aligned with image dimensions, enabling a more nuanced assessment of feature quality. Inspired by this literature, our intuition is to leverage such probabilities as direct feedback to assess image utility.




To quantify image utility, we utilize R2D2 as the base feature detection network. We compute the product of the mean repeatability and the square of the mean reliability that yields a single scalar performance index \(M_{feat}\) per image: 
\begin{equation}
\label{eq:mfeat}
M_{\text{feat}} = \left( \frac{1}{N} \sum_{i=1}^{N} R_i \right) \cdot \left( \frac{1}{N} \sum_{i=1}^{N} Q_i \right)^2
\end{equation}

This metric describes the image's utility for successful feature detection. The square of the mean reliability amplifies its influence on the final score, reflecting the critical role of the descriptor in accurate matching performance. Here, \(R_{i}\) is repeatability per pixel, \(Q_{i}\) is reliability per pixel, and \(N\) is the total number of pixels.
We evaluated \(M_{feat}\)'s performance and compared it against other metrics in \autoref{Experiments_and_Results} where \(M_{feat}\) shows a strong correlation with feature matching performance.

\subsection{Illumination and Exposure Control}



The illumination and exposure control problem is formulated as a multi-variable Bayesian optimization (BO) where the optimal value of the light-intensity (\(P\)) and exposure-time (\(\Delta t\)) are determined by maximizing our image quality metric $M_{feat}$.
The Gaussian process (GP) provides the surrogate model to parameterize the influence of $P$ and $\Delta t$ with zero mean Gaussian noise $n = \mathcal{N}(0,\sigma_n^2)$. 
The mean $\mu$ and covariance $\sigma^2$ of the predictive distribution is given by:
\begin{gather}
   \mu(\textbf{x}_*) = k_*^T (K + \sigma_n^2 I)^{-1} y  \\
   \sigma^2(\textbf{x}_*) = k(\textbf{x}_*, \textbf{x}_*) - k^T_* (K + \sigma_n^2 I)^{-1} k_* 
\end{gather} 
where $\textbf{x}$ = [$P$,$\Delta T$] gives inputs $\textbf{x}_* = \{\textbf{x}_0, \dots,\textbf{x}_m\}$ and outputs $y =\{y_0, \dots,y_m\}$. In this work, the Mat\'{e}rn kernel \cite{rasmussen_gaussian_2006} function is selected to construct the correlation matrix $K$ and vector $k$. In addition, Expected Improvement~\cite{snoek_practical_2012} (EI)  is selected as an acquisition function to determine the next evaluation points.
\begin{figure}
    \centering
    \includegraphics[width=\linewidth, trim=0.5cm 1.0cm 0.8cm 0cm, clip]{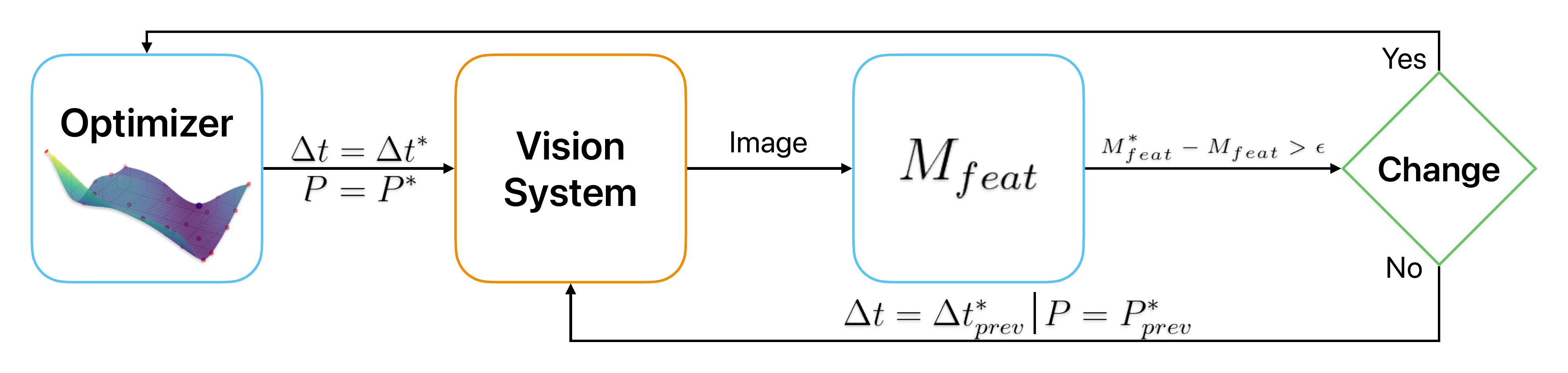}
    \caption{Overview of the \qq{} pipeline}
    \label{fig:arch}
\end{figure}

\qq{}'s overall architecture is illustrated in \autoref{fig:arch} where the algorithm begins with BO to compute the optimal configuration (\(P^*\) and  \(\Delta t^*\)) which 
provides the optimal $M_{feat}^*$. After applying the configuration, images are received by an image quality assessment module which checks the current metric value and compares it with the optimal. A threshold \(\epsilon\) ($\epsilon >0$) is provided by the user as the tolerance.  As the robot moves, if $M_{feat}^* - M_{feat} > \epsilon$,  the system triggers another round of optimization. 
\section{Experiments and Results}
\label{Experiments_and_Results}

\begin{figure}
\vspace{-0.2cm}
    \centering
    \includegraphics[width=0.8\linewidth, trim=0.5cm 1.4cm 1.2cm 0cm, clip]{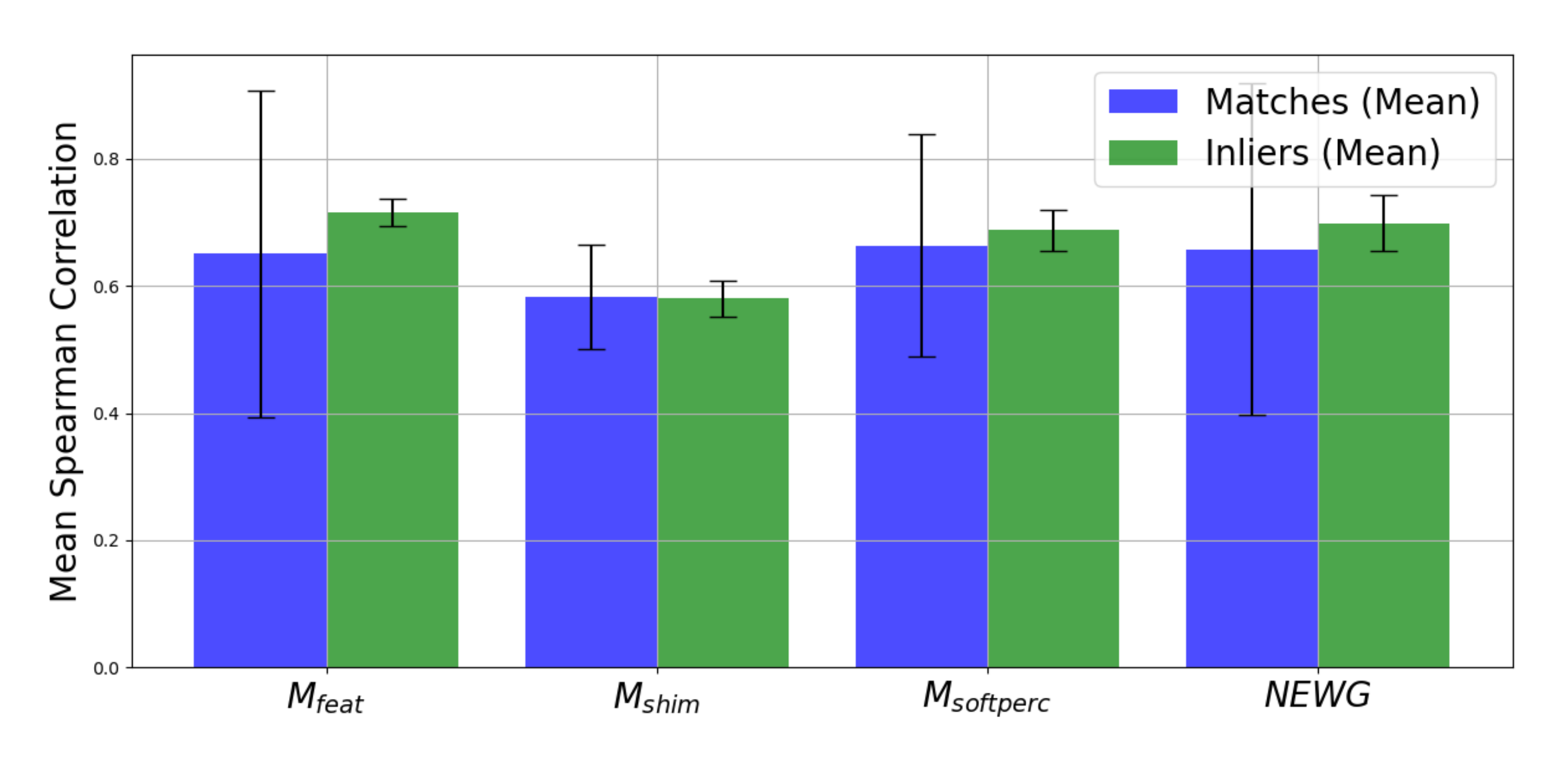}
    \caption{Correlation of utility metrics with feature detection/matching. }
    \yash{Reviewer :  NEWG matches error-bar not visible}
    \label{fig:bar-combo}
\end{figure}

\RED{We benchmark \(M_{feat}\) against established utility measures such as \(M_{shim}\), \(NEWG\)\, and \(M_{softperc}\). }
\autoref{fig:bar-combo} shows the correlation between the metrics and feature matching performance between consecutive frames. Spearman correlation is used to quantify the relationship between each metric and feature matching. 



\begin{figure}
    \centering
    \includegraphics[width=\linewidth, trim=0.5cm 1.4cm 1.2cm 0cm, clip]{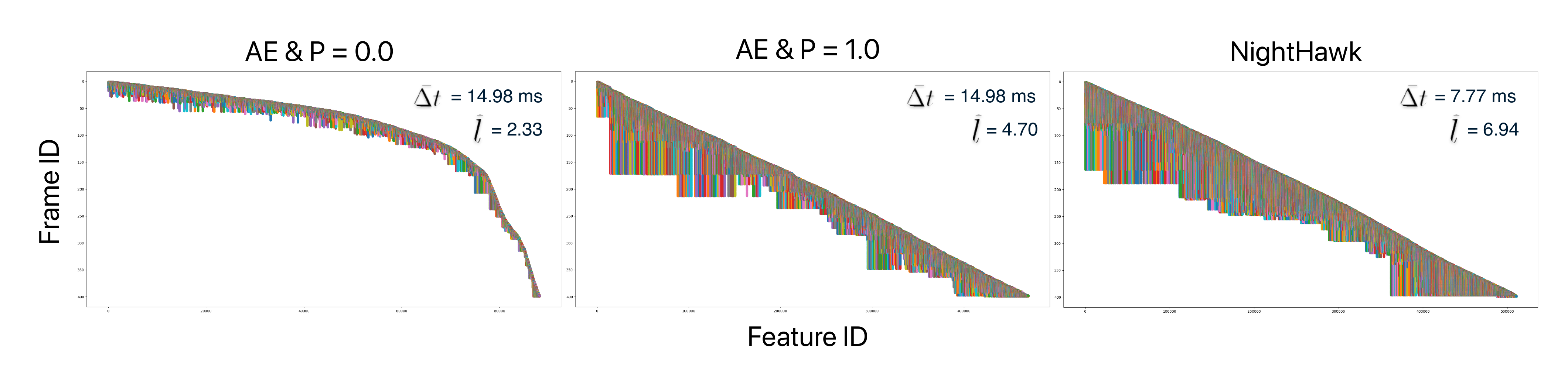}
    \caption{Feature tracking performance of the 3 settings where \qq{} shows improved feature tracking and lower exposure-times}
    \label{fig:feat-tracks}
\end{figure}
\(M_{feat}\) demonstrates a strong positive correlation with feature matching performance across five diverse feature detectors (AKAZE \cite{alcantarilla_fast_2013}, SHI\_TOMASI \cite{jianbo_shi_good_1994}, ORB \cite{rublee_orb_2011}, R2D2, and Superpoint \cite{detone_superpoint_2018}). While \(NEWG\) also exhibits comparable correlation, \(M_{feat}\) is more consistent with  lower variance. Further, \(M_{feat}\)'s capacity to incorporate information from learning-based feature detectors distinguishes it from conventional metrics reliant solely on gradient-based features and enables a more comprehensive evaluation of image utility.
Finally, \(M_{feat}\) offers a significant practical advantage when using learning-based features in visual estimation. A single computation for both feature extraction and quality assessment (e.g., when using R2D2 features) can help reduce compute overheads.


We deploy \qq{} on a Boston Dynamic's Spot robot equipped with a FLIR Blackfly S camera and controllable 50W LED.
\qq{} was implemented using ROS2 \cite{macenski_robot_2022}, PyTorch \cite{paszke_pytorch_nodate}, and Scikit-learn \cite{pedregosa_scikit-learn_nodate} and runs online on an onboard NVIDIA Jetson Orin. We use a pre-trained R2D2 model provided by the authors for our \(M_{feat}\) image utility metric. The optimization process \RED{duration can be as short as 20 seconds} depending on the chosen hyper-parameters. We conducted several tele-operated missions beneath the Erie Canal, capturing images under 3 camera configurations with fixed gain: 



\begin{enumerate}
    \item  \textbf{Auto-exposure \(+\) no external light (AE \& P=0.0)}: The camera relied solely on its built-in AE mechanism.
    \item  \textbf{Auto-exposure \(+\) fixed external light (AE \& P=1.0)}:  The camera used AE while a constant 100\% intensity from the LED was applied.
    \item  \textbf{\qq{} optimization}: Our proposed method dynamically adjusted both exposure-time (\(\Delta t\)) and LED intensity (\(P\)) to optimize image quality.
\end{enumerate}

\begin{figure}
    \centering
    \includegraphics[width=1\linewidth, trim=1.8cm 1.5cm 1.8cm 1.0cm, clip]{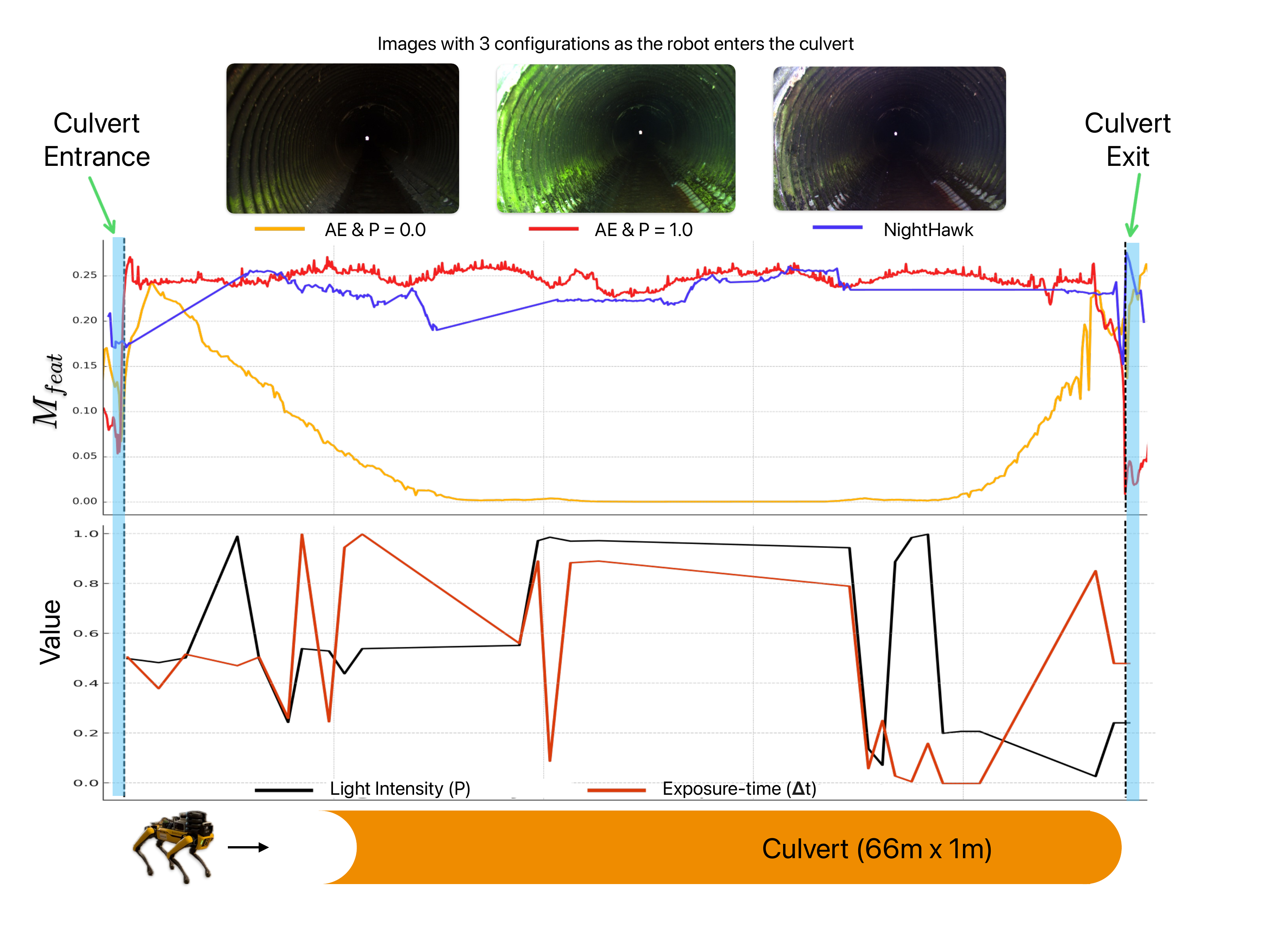}
    \caption{Change in \(M_{feat}\) as the robot enters the culvert is shown in the 3 configurations, AE with no external light results in underexposed images while adding a fixed (100\%) light introduces artifacts such as a green-ish hue. Finally, with \qq{}, overall image utility is consistent, the sudden dips in \(M_{feat}\) are prevented and exposure-times and light intensity is well-balanced.} 
    \label{fig:score-combo}
\end{figure}

\RED{
\autoref{fig:score-combo} shows the image utility \(M_{feat}\) as robot navigates through the culvert environment while \autoref{fig:feat-tracks} shows feature tracking performance. The highlighted regions indicate the time intervals during which the robot enters and exits the culvert.
When using (AE \& P = 0.0) configuration, \(M_{feat}\) drops sharply as the robot enters the culvert. Although AE initially adjusts to improve the image quality, \(M_{feat}\) eventually falls to near zero due to the environment's darkness and AE's limitations at maximum exposure settings.
In the (AE \& P = 1.0) configuration, a similar initial drop in \(M_{feat}\) occurs as the robot transitions into the culvert. This is likely caused by reflections and exposure instability when the lighting suddenly changes. AE recovers after this initial drop, but the inconsistency as a lasting impact on feature detection and matching.
With \qq{}, however, \(M_{feat}\) exhibits a lower value compared to fixed light but avoids sharp drops. This results in more stable image utility throughout the transition, supporting more consistent feature detection and matching. A similar trend is observed when the robot exits the culvert.

}


Feature tracking performance, as shown in \autoref{fig:feat-tracks}, is sub-optimal for the two baseline methods. In the \textit{AE \& P = 0.0} setting, limited feature persistence ($\bar{l} = 2.33$) results in short track lengths and poor image utility, despite a standard exposure time ($\Delta t = 14.98$ ms). Adding external light in the \textit{AE \& P = 1.0} setting improves the average track length ($\bar{l} = 4.70$), but introduces visual artifacts such as a green-ish hue due to reflections and uneven illumination. In contrast, \textit{NightHawk} achieves the best performance, with significantly higher average track length ($\bar{l} = 6.94$) and reduced exposure time ($\Delta t = 7.77$ ms), enabling longer and more consistent feature tracking while minimizing motion blur and avoiding overexposure.




\RED{
Beyond enhancing feature performance, \qq{} also maintains well balanced exposure times and light intensities by optimizing camera settings within a user-defined search space. Balanced exposure not only ensures consistent image quality but also contributes to reduced power consumption by avoiding unnecessary use of external lighting. Since camera exposure time directly limits the achievable frame rate, we constrain the maximum allowable exposure time in the optimizer based on the desired frame rate. This guarantees that \qq{} produces solutions that meet the real-time performance requirements of our application while operating efficiently in terms of both computation and energy usage.}


\begin{figure}
    \centering
    \includegraphics[width=0.9\linewidth,trim=1.8cm 1.5cm 0.8cm 1.0cm, clip]{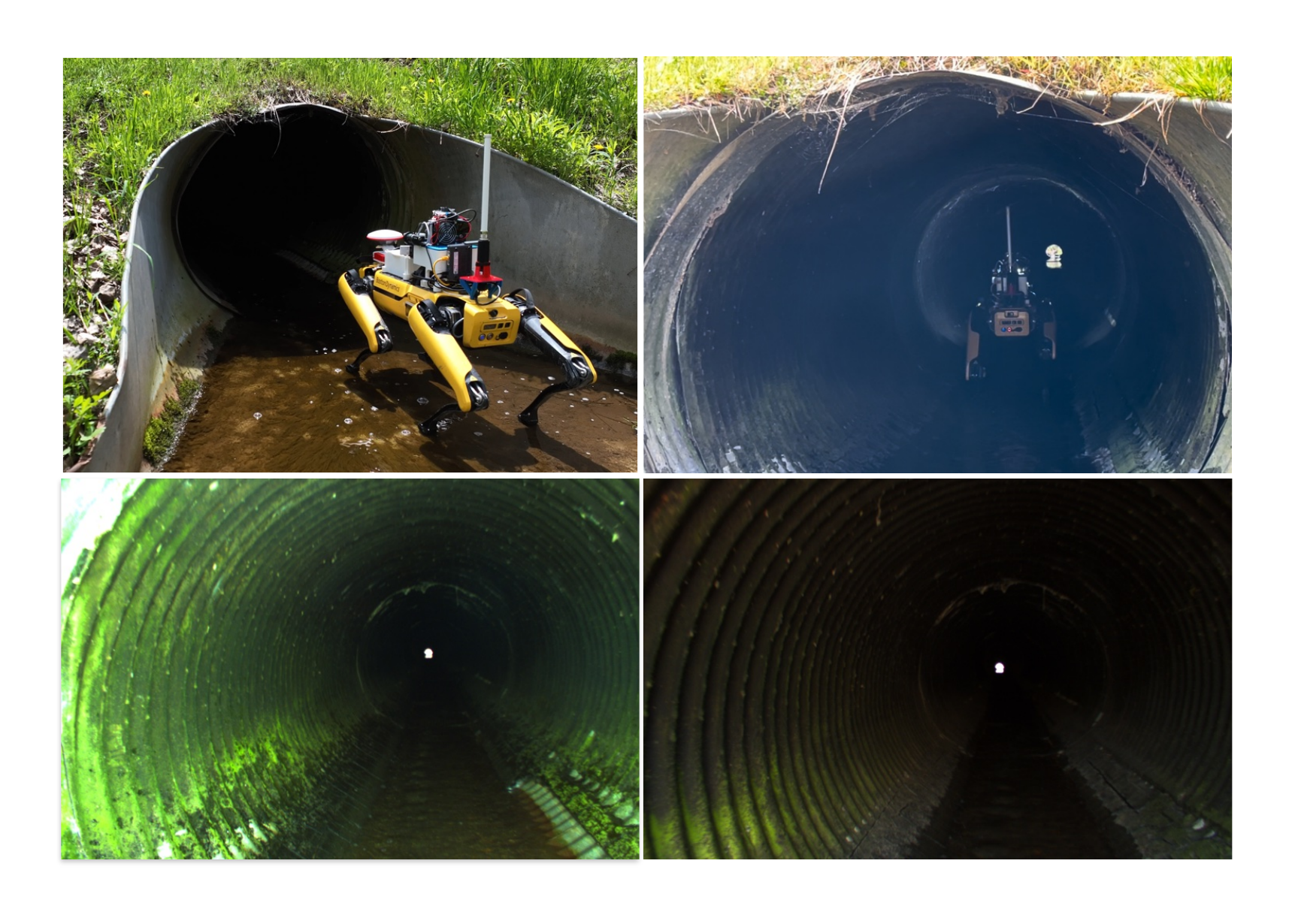}
    \caption{
    \yash{Fix / update caption}
    Deployment of our active illumination system in subterranean culvert environments.  The top row shows the robot equipped with \qq{} approaching and entering a culvert beneath the Erie Canal. The bottom row illustrates the visual challenges encountered inside the culvert, including low-light conditions, overexposure, and artifacts from onboard lighting. Notably, the left bottom image exhibits a strong green hue caused by the interaction of fixed illumination with reflective surfaces and wet tunnel walls. These conditions motivate the need for adaptive, scene-aware illumination control to improve image quality and feature visibility.}
    \label{fig:insights-tab}
    \vspace{-0.8cm}
\end{figure}
\section{Experimental Insights}
\label{Experimental_Insights}

To evaluate the performance of \qq{} in real-world scenarios, we conducted field experiments in subterranean environments. These trials, illustrated in Figure~\ref{fig:insights-tab}, highlight the challenges posed by low-light, reflective, and texture-sparse conditions, and demonstrate how adaptive exposure and illumination significantly improve visual perception.

\subsection{Inspection in Low Light Conditions}

{\it Optimizing Adaptive Lighting}: A key challenge in low-light and dynamic lighting environments for robot perception is the ability to adapt. Our observation is that controlling only exposure settings of a camera, or mounting a light that is switched on throughout the task gives suboptimal feature detection and tracking results. 
\begin{figure}
    \centering
    \includegraphics[width=1\linewidth]{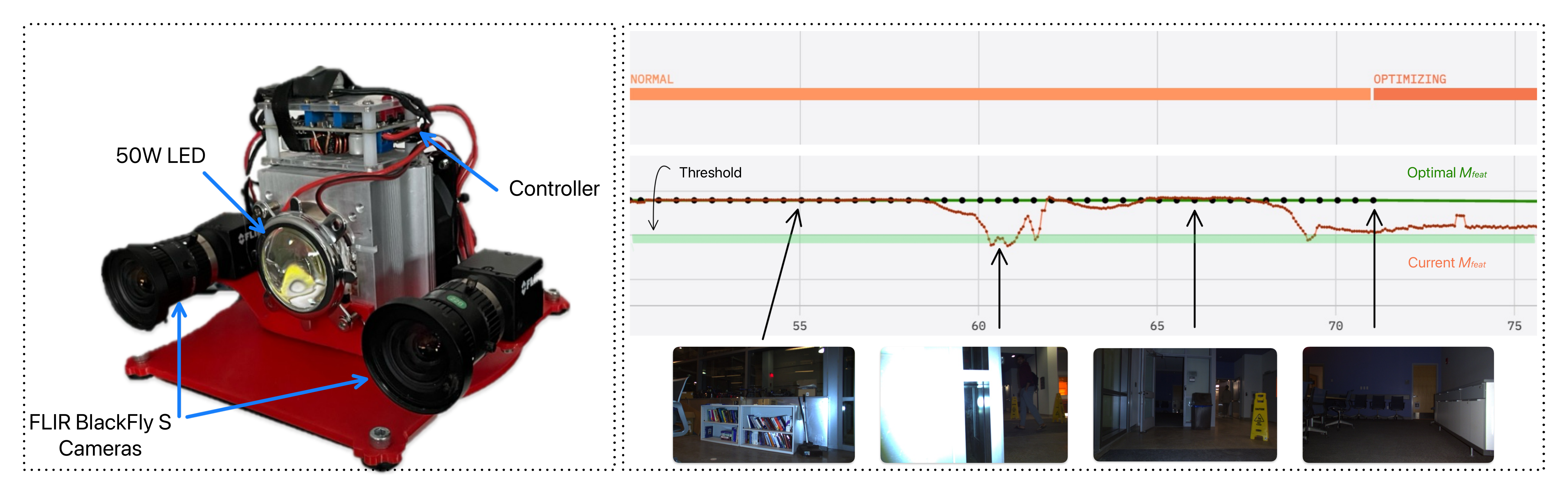}
    \caption{\qq{} system in action. (Left)  Hardware prototype featuring a high-power 50W LED, stereo FLIR BlackFly S cameras, and an onboard ROS-based LED controller. (Right) Real-time response of the system during a live experiment, where \(M_{feat}\) drops below a threshold, triggering the optimization routine. As the system adapts, the metric improves, demonstrating \qq{}’s ability to respond to challenging lighting conditions and recover usable visual data. Notably, \qq{} is robust to short-duration drops in \(M_{feat}\); optimization is only triggered if the threshold is violated for a set number of consecutive frames. This design prevents unnecessary reconfiguration in response to ephemeral lighting effects. Sample images illustrate transitions and optimized states.}
    \label{fig:impl-dets}
\end{figure}

Controlling an adjustable light, identifying when the lighting conditions have changed, and adaptively reconfiguring the light as well as exposure jointly leads to much better feature detection/tracking (\autoref{fig:feat-tracks}) including a 47\% improvement in feature tracking in our scenario. This also results in more robust robot perception as shown by the low variance in correlation in \autoref{fig:bar-combo}. 

{\it Tuning to a given scenario: }There are several tradeoffs in incorporating \qq{} into an inspection system. The optimization takes time as seen by the delays in execution (several seconds per run) once the optimization is triggered in \autoref{fig:impl-dets}. Additionally, we need to tune the threshold ($\epsilon$) to ensure the optimization is not triggered too frequently to affect task execution while also not triggering too infrequently to affect useful image capture. In our executions, we have tuned the threshold to adjust to big changes - once when it goes from the outside into the culvert, and again when it is deep enough in the culvert that the lighting has completely changed. Note that \qq{} changes the light setting to 54\% intensity when it enters the culvert, far less than the 100\% setting in our baseline. This results in slightly lower scores as in \autoref{fig:score-combo}, but good feature matching accuracy in our application.

{\it Integrating other perception services: }In theory, we can increase exposure unboundedly to improve image capture. However, in most robot autonomy, there are other services such as localization and mapping that assume continuity through motion and a certain frame rate for efficient execution \cite{semenova_comprehensive_2025},\cite{semenova_quantitative_2022}. Further, motion adds blur which is exacerbated by long exposure. We observe that adaptive lighting allows us to function at reasonable frame rate to cater to such services while improving inspection in the culvert environment.

{\it Energy Efficiency: }A side benefit is reduced energy expenditure. Our setup in \autoref{fig:impl-dets} has a 50W LED. The total power output of the robot is 150W. Keeping a light powered on at full intensity can severely limit the robot range and endurance. \RED{This is validated in \autoref{fig:score-combo} which shows the light intensity vary as the robot navigates the culvert.}


\subsection{Learning-based Metric } 

Our metric builds on ongoing research in learning-based feature detectors, and uses R2D2. The advantage of using our metric is that we can both compute the image utility as well as detect features in one pass. As demonstrated in \autoref{fig:bar-combo}, it exhibits strong and reliable (low variance) correlation with feature detection and matching. Beyond our application, we believe that our metric can be widely applicable for applications requiring to only adjust auto-exposure, or even evaluate the image utility of a given scene for other purposes. 
\subsection{System Optimizations}

{\it Onboard Optimization:} Optimization must efficiently execute on resource constrained robot hardware. We employ multi-threading and early stopping to accelerate convergence. Through this process, we reduced the latency of optimization from 70 seconds to 20 seconds.

{\it Parameter Control and Feedback:} The optimizer adjusts  \(P\), \(\Delta t\), captures images, and receives feedback \(M_{feat}\). ROS2 ensures time synchronization and precise hardware control.

{\it Utility Computation:} \(M_{feat}\) is computed online at high rates by applying CUDA acceleration to down-sampled images, achieving 15Hz on an NVIDIA Jetson Orin while running other tasks on the CPU. 

{\it Event-triggered Maneuvers:} The optimizer can temporarily pause robot movement, determining optimal settings before proceeding to the next way-point. One promising solutions is to move this evaluation to the background in an predictive manner, which will significantly reduce the delay.
\section{Conclusion and Future Plan}

We propose \qq{}, an active illumination and exposure control method that enhances visual estimation in low-light and dynamically lit environments. Field experiments demonstrate that deploying \qq{} in challenging environments improves feature tracking performance by 47--197\%. These experiments also provide valuable insights into how exposure settings and onboard lighting affect feature detection, offering a foundation for applications where low-light visual estimation is critical.

In future work, we aim to generalize the Bayesian optimization process using learning-based methods, allowing \qq{} to perform exposure and illumination control in a single shot. This advancement would eliminate the need for the current event-triggered (stop-and-go) routine, significantly accelerating navigation.

\section{Acknowledgments}

We thank our colleagues Christo Aluckal, Kartikeya Singh, and Yashom Dighe for their invaluable assistance with the experiments. We also extend our gratitude to Matthew Lengel from the NY Canal Corporation for providing access to inspection sites and sharing his expertise on culvert infrastructure. 

%
%
%


\bibliographystyle{plain}
%
\bibliography{cleaned_refs} 





\end{document}